\documentclass{article}
\usepackage{ijcai16}
\usepackage{times}
\usepackage{graphbox}
\usepackage{multirow}
\usepackage{placeins}
\usepackage[font=small]{caption}

\RequirePackage{amsmath,amssymb,amsfonts,amsthm}
\usepackage{algorithm}
\usepackage{algpseudocode}

\algrenewcommand{\algorithmicrequire}{\textbf{Input:}}
\algrenewcommand{\algorithmicensure}{\textbf{Output:}}

\renewcommand\[{\begin{equation}}
\renewcommand\]{\end{equation}}

\newcommand{\bbR}{\mathbb{R}}

\newcommand{\calvar}[1]{\ensuremath{\mathcal{#1}}}

\newcommand{\calD}{\calvar{D}}

\newcommand{\calX}{\calvar{X}}

\newcommand{\vecvar}[1]{\ensuremath{\boldsymbol{#1}}}

\newcommand{\vv}{\vecvar{v}}
\newcommand{\vw}{\vecvar{w}}
\newcommand{\vx}{\vecvar{x}}
\newcommand{\vy}{\vecvar{y}}

\newcommand{\veps}{\vecvar{\varepsilon}}

\usepackage{color}

\newcommand{\url}[1]{#1}
\title{Constructive Preference Elicitation by Setwise Max-margin Learning}

\author{Stefano Teso \\
University of Trento \\
Trento, Italy \\
teso@disi.unitn.it
\And
Andrea Passerini \\
University of Trento \\
Trento, Italy \\
passerini@disi.unitn.it
\And
Paolo Viappiani \\
Sorbonne Universit\'es\\
UPMC Univ Paris 06\\
CNRS, LIP6 UMR 7606\\
Paris, France \\
paolo.viappiani@lip6.fr}

\begin{document}

\maketitle

\begin{abstract}
  In this paper we propose an approach to preference elicitation that
  is suitable to large configuration spaces beyond the reach of
  existing state-of-the-art approaches. Our setwise max-margin method
  can be viewed as a generalization of max-margin learning to sets,
  and can produce a set of ``diverse'' items that can be used to ask
  informative queries to the user.  Moreover, the approach can
  encourage sparsity in the parameter space, in order to favor the
  assessment of utility towards combinations of weights that
  concentrate on just few features.  We present a mixed integer linear
  programming formulation and show how our approach compares
  favourably with Bayesian preference elicitation alternatives and
  easily scales to realistic datasets.
\end{abstract}

\section{Introduction}
Preferences \cite{Peintner2008} play an important role in a variety of artificial
intelligence applications and the task of eliciting or learning preferences is a crucial one; 
typically only limited
information about the user's preferences will be available and the
cost (cognitive or computational) of obtaining additional preference information will be high.  
%The goal in
%preference elicitation is that of gaining enough information about the
%user's
%preferences %(typically represented by an utility function $u$, unknown to the system)
%in order to recommend a decision or a course of action.

The automated assessment of preferences has received considerable attention,
starting with pioneering works in the OR community, such as \cite{White1984}
and especially the UTA methodology \cite{jaquetlsiskos1982} giving rise to a
wide variety of extensions \cite{jaquetlsiskos2001,greco2008ordinal}.
%In the past decade a number of researchers have considered elicitation of preferences from an artificial intelligence perspective%; within the machine learning field the term preference learning has been catching up.
%While preferences are often assessed using some ad-hoc rules \cite{},
Within AI, a number researchers have proposed interactive methods that elicit
preferences in an adaptive way
\cite{chajewska2000,boutilier2002,Wang2003,boutilier2006,guo2010real,viappiani2010optimal},
observing that, by asking informative questions, it is often possible to make
near-optimal decisions with only partial preference information.

%focusing on learning the ``important'' part of the
%preferences, 
%braziunas-mmr:uai07,

%One option is to be Bayesian and to maintain a probability distribution over the possible realizations of the utility's parameters \cite{chajewska2000,boutilier2002,guo2010real,viappiani2010optimal}.
%Alternatively, when distribution information about the parameters is not available, one can reason about all feasible utility functions consistent with the currently known information about the user; from the user's response we are able to infer constraints on the parameters implicated in the utility model \cite{Wang2003,boutilier2006,braziunasmmr:uai07,viappiani2009}.
%This latter approach has also the advantage of avoiding the computationally demanding task of maintaining distribution information and performing Bayesian updates.

% constructive preference elicitation
While most works assume that items or decisions are available in a (possibly large)
dataset, in this paper we propose an adaptive elicitation framework that takes a {\em constructive} view on preference
elicitation, enlarging its scope from the selection of items among a
set of candidates to the synthesis of entirely novel instances. 
Instances are solutions to a given optimization problem; they are represented as combinations of basic elements
(e.g. the components of a laptop) subject to a set of constraints
(e.g. the laptop model determines the set of available CPUs). 
A utility function is learned over the feature representation of an
instance, as customary in many preference elicitation approaches. 
The recommendation is then made by solving a constrained optimization
 problem in the space of feasible instances, guided by the learned
utility. 

Preference elicitation in configuration problems has been previously
tackled with regret-based elicitation
\cite{boutilier2006,braziunas2007}, where minimax regret is used both
as a robust recommendation criterion and as a technique to drive
elicitation.  The main limitation of their
approach is the lack of tolerance with respect to user
inconsistency. %(the user might state that an item is preferred to
%another one, even though the latter has higher utility).
% setwise max-margin formulation to deal with uncertainty in user utility
Indeed, learning a user utility function requires 
%elicitation strategy and 
to deal with uncertain and possibly inconsistent user feedback.  

Bayesian preference elicitation
approaches deal with this problem by 
building a probability
distribution on candidate functions (endowed with a response or error
model to be used for inference) and asking queries maximizing
informativeness measures such as {\em expected value of information
(EVOI)}~\cite{chajewska2000,guo2010real,viappiani2010optimal}.  
These
approaches are however computationally expensive and can not scale to
fully constructive scenarios, as shown in our experimental results.

We take a space decomposition perspective and jointly learn a set of
weight vectors, each representing a candidate utility function,
maximizing diversity between the vectors and consistency with the
available feedback. These two conflicting objectives tend to generate
equally plausible alternative hypotheses for the unknown
utility. 
Our approach to elicitation works by combining weight vector
learning with instance generation, so that each iteration of the
algorithm produces two outcomes: a set of weight vectors and a set of
instances, each maximizing its score according to one of the weight
vectors. 
We evaluate the effectiveness of our approach by testing our
elicitation method in both synthetic and real-world problems, and
comparing it to state-of-the-art methods.

% The paper is structured as follows: we present some background
% material in Section \ref{sec:background} and then proceed to our
% setwise max-margin method in Section \ref{sec:formulation}; we discuss
% experimental results in Section \ref{sec:experiments} and conclude
% with final remarks (Sec.~\ref{sec:conclusions}).

%Preference elicitation for the customization of user interfaces \cite{gajos2005}.

%WRITEME

\section{Background}
\label{sec:background}
%\section{Preference Elicitation}

%\paragraph{Notation.}
We first introduce some notation.
We use boldface letters $\vx$ to indicate vectors,
uppercase letters $X$ for matrices, and calligraphic capital letters $\calX$
for sets. We %frequently 
abbreviate the set $\{ x^i \}_{i=1}^n$ as $\{ x^i \}$
whenever the range of the index $i$ is clear from the context, and use $[n]$ as
a shorthand for %the set 
$\{1, \ldots, n\}$. We write $\|\vx\|_1 := \sum_z |x_z|$
to indicate the $\ell_1$ vector norm, $\langle \cdot, \cdot \rangle$ for the usual dot product,  $X'$ for matrix transposition.

%\paragraph{Setting.} 
We assume to have a multi-attribute feature space
$\calX$ of configurations $\vx = (x_1, \ldots, x_m)$ over $m$
features. For the sake of simplicity we focus on binary features only,
i.e. $x_z\in\{0,1\}$ for all $z\in[m]$, assuming a one-hot encoding of
categorical features. This is a common choice for preference
elicitation methods~\cite{guo2010real,viappiani2010optimal}. Support
for linearly dependent continuous features will be discussed later on.

We further assume that the set of {\em feasible} configurations, denoted by
$\calX_\text{feasible}$, is expressed as
a conjunction of linear constraints. This allows to formulate both arithmetic
and logical constraints,
e.g. under the canonical mapping of $True$ to $1$ and $False$ to $0$, the
Boolean disjunction of two binary variables $x_1 \lor x_2$ can be rewritten as
$x_1 + x_2 \ge 1$.

Consistently with the experimental settings of previous work~\cite{guo2010real,viappiani2010optimal},
we model users with additive utility functions \cite{keeney1976}; 
the user's preferences are represented by a  weight
vector $\vw\in\bbR^m$ and
%We assume linear utility functions \cite{keeney1976}; 
the utility of a configuration $\vx$ is given by
$\langle \vw, \vx \rangle = \sum_{z=1}^m w_z x_z$. 
In the remainder of the paper we
require all weights to be {\em non-negative} and {\em bounded}: the per-attribute
weights $w_z$ must lie in a (constant but otherwise arbitrary) interval
$[w^\bot_z, w^\top_z]$, with $w^\bot_z \ge 0$. 
Both requirements are quite natural\footnote{Utility values are defined on an
interval scale, thus it is always possible to scale the values appropriately (see for instance \cite{Torra2007} and \cite{keeney1976}).}, and enable the translation of our core optimization problem into a
mixed-integer linear problem (as done in Section~\ref{sec:formulation}). 

During learning, the actual weight vector $\vw$ is {\em unknown} to
the learning system, and must be estimated by interacting with the
user. We mostly focus on pairwise comparison queries, which are the
simplest of the comparative queries. These can be extended to choice
sets of more than two options
\cite{viappiani2009,viappiani2010optimal} and are common in conjoint analysis
\cite{louviere2000,toubia2004}. %We consider three possible outcomes
For a pairwise comparison between two configurations $\vx$ and $\vx'$:
either $\vx$ is preferred to $\vx'$ (written $\vx \succ \vx'$), $\vx'$
is preferred to $\vx$ ($\vx \prec \vx'$), or there is no clear
preference between the two items ($\vx \approx \vx'$). We write
$\calD$ to denote the set of preferences (answers to comparison
queries) elicited from the user. 

In the next Section we describe how informative queries can be generated using our setwise maxmargin learning.

\section{Setwise Max-margin Learning}
\label{sec:formulation}
\paragraph{Non-linear Formulation.} We first introduce the problem
formulation as a %mixed integer 
non-linear optimization problem, and
then show how to reduce it to a mixed integer linear program.% optimization one.

The goal of our setwise max-margin approach is twofold. First, for any
given set size $k\geq 1$, we want to find a {\em set} of $k$ weight
vectors $\vw^{1}, \ldots, \vw^{k}$, chosen so that all user-provided
preferences are satisfied by the largest possible margin (%
%i.e. all
%weight vectors are consistent with respect to the user responses
%$\calD$, 
modulo inconsistencies) and so that they are maximally
diverse.  Second, we want to construct a {\em set} of $k$
configurations $\vx^{1}, \ldots, \vx^{k}$, so that each configuration
$\vx^{i}$ is the ``best'' possible option when evaluated according to
the corresponding $\vw^{i}$ {\em and} configurations are maximally diverse
among each other. These options will be later used to formulate
queries.

The first goal is achieved by translating all pairwise preferences
$\calD$ into ranking constraints: preferences of the form
$\vy^h_+ \succ \vy^h_-$ become linear inequalities of the form
$\langle \vw^i, \vy^h_+ - \vy^h_- \rangle \geq \mu$, where $\mu$ is the
{\em margin} variable (which we aim at maximizing) and $h$ ranges over
the responses.  Non-separable datasets, which occur in practice due to
occasional inconsistencies in user feedback, are handled by
introducing slack variables (whose sum we aim at minimizing)
in a way similar to UTA and its extensions \cite{jaquetlsiskos1982,greco2008ordinal}. When
augmented with the slacks, the above inequalities take the form
$\langle \vw^{i}, \vy^{h}_+ - \vy^{h}_- \rangle \ge \mu - \varepsilon^{i}_h$
where $\varepsilon^{i}_h$ is the penalty incurred by weight vector $\vw^{i}$
for violating the margin separation of pair $h$. Indifference preferences, i.e.
$\vy^h_1 \approx \vy^h_2$, are translated as $|\langle \vw^i, \vy^h_1 - \vy^h_2 \rangle| < \varepsilon^i_h$;
the slack increases with the difference between the estimated utility of the
two options.

The second goal requires to jointly maximize the utility of each
$\vx^{i}$ according to its corresponding weight vector $\vw^i$ and its
scoring difference with respect to the other configurations $\vx^j$ in
the set. We achieve this by maximizing the sum of utilities
$\sum_{i=1}^k \langle \vw^{i}, \vx^{i} \rangle$ and adding ranking
constraints of the form
$\langle \vw^{i}, \vx^{i} - \vx^{j} \rangle \geq \mu$ for all
$i,j\in[k]$, $i \ne j$.

% Ideally, the second goal would be implemented as:
% %
% $$ \vx^{i} = \argmax_{\vx \in \calX_\text{feasible}} \langle \vw^{i}, \vx^{i} \rangle $$
% %
% However this formulation is highly impractical. We therefore only require 
% each option $\vx^{i}$ to be the best {\em among} the configurations
% $\{ \vx^i \}$. This can be accomplished by imposing
% constraints of the form $\langle \vw^{i}, \vx^{i} - \vx^{j} \rangle \geq \mu$
% for all $i,j\in[k]$, $i \ne j$.
% This however does not guarantee that the produced $\{ \vx^{i} \}$ have a high
% {\em absolute} utility. We therefore favor high-quality configurations by
% introducing an additional term $\sum_{i=1}^k \langle \vw^{i}, \vx^{i} \rangle$
% in the objective function.

A straightforward encoding of the above desiderata leads to the
following mixed integer {\em non-linear} optimization problem over the
non-negative margin $\mu \in \bbR_{\ge 0}$ and vectors $\{ \vw^i \in \bbR^m \}$, $\{ \vx^i \in \{0,1\}^m \}$:
{\footnotesize
\begin{align}
    \max
        & \;\; \mu - \alpha \sum_{i=1}^k \| \veps^{i} \|_1 - \beta \sum_{i=1}^k \| \vw^{i} \|_1 + \gamma \sum_{i=1}^k \langle \vw^{i}, \vx^{i} \rangle
        \nonumber
    \\
    \text{s.t.}
        & \;\; \forall \; i \in [k], \forall \; h \in [n] \quad \langle \vw^{i}, \vy^{h}_+ - \vy^{h}_- \rangle \ge \mu - \varepsilon^{i}_h \label{eq:wyconstr}
    \\
        & \;\; \forall \; i, j \in [k], i \neq j \quad \langle \vw^{i}, \vx^{i} - \vx^{j} \rangle \ge \mu \label{eq:wxconstr}
    \\
        & \;\; \forall \; i \in [k] \quad \vw^\bot \le \vw^{i} \le \vw^\top \label{eq:wbounds}
    \\
        & \;\; \forall \; i \in [k] \quad \vx^{i} \in \calX_{\text{feasible}} \;\;, \veps^{i} \ge 0 \label{eq:xbounds}
%    \\
%        & \;\; \forall \; i \in [k] \quad \vx^{i} \in \calX_{\text{feasible}} \label{eq:xbounds}
%    \\
%        & \;\; \forall \; i \in [k] \quad \veps^{i} \ge 0 \nonumber
%    \\
%        & \;\; \mu \ge 0 \nonumber
\end{align}
}
Let us illustrate the above piece by piece. The objective is composed
of four parts: we maximize the shared margin $\mu$ (first part) and
minimize the total sum of the ranking errors $\veps^i$ incurred by
each weight vector $\vw^{i}$ (second part), while at the same time
regularizing the magnitude of the weights (third part) and the quality
of the configurations $\{ \vx^{i} \}$ (last part). The non-negative
hyperparameters $\alpha,\beta,\gamma$ control the influence of the
various components. The weight regularization term copes with the
common scenario in which the user has strong preferences about some
attributes, but is indifferent to most of
them. %~\cite{} \stefano{@Paolo: add refs}.
The $\ell_1$ penalty is frequently used to improve the sparsity of
learned models~\cite{lasso,zhang2008,Hensinger2010}, with consequent
gains in generalization ability and efficiency, as confirmed by our
empirical findings (see
Section~\ref{sec:experiments}). Constraint~(\ref{eq:wyconstr})
enforces the correct ranking of the observed user preferences, while
%constraint
~(\ref{eq:wxconstr}) ensures that the generated
configurations are diverse in terms of the weight vectors they
maximize. Constraints~(\ref{eq:wbounds}) and~(\ref{eq:xbounds}) ensure
that the weights and configurations are feasible
%, while the last one
and
guarantees the non-negativity of the slacks.  Since we require
$\vw^\bot \ge (0,\ldots,0)$, Eq.~(\ref{eq:wbounds}) also enforces the
weights to be non-negative.

\begin{figure}[t]
    \begin{center}
        \includegraphics[width=10em]{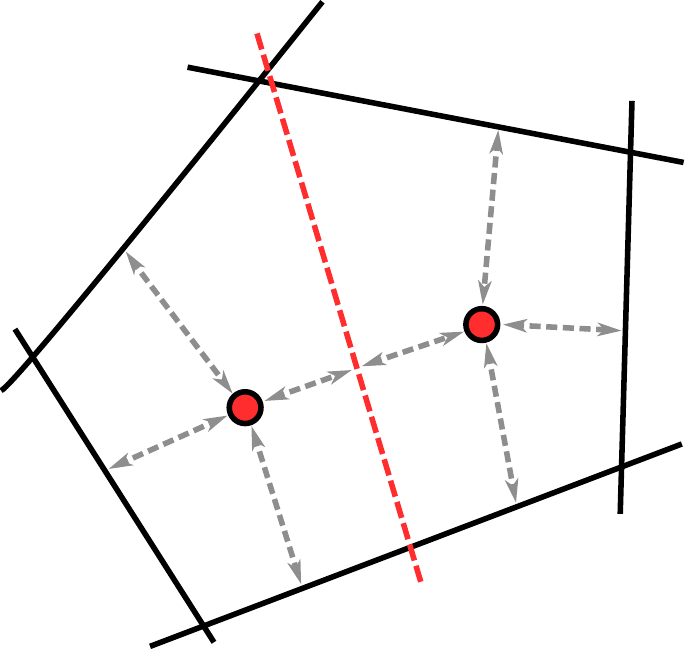}
    \end{center}
    \caption{\label{fig:setmargin} Optimization of setwise max-margin; the black lines corresponds to preference constraints $\calD$, the red points are utility vectors $\vw^{1}$ and $\vw^{2}$, the red line corresponds to the hyperplane $\langle \vw, \vx^{1} - \vx^{2} \rangle = 0$.}
\end{figure}

Note that we are choosing the configurations $\{ \vx^i \}$ and the
weight vectors $\{ \vw^i \}$ {\em simultaneously}.  
%Note that the methods bears some similarity with volumetric approaches, but  with important differences. 
We look for $\vw^i$ so that the utility loss (see constraint \ref{eq:wxconstr}) of choosing $\vx^{j}$ instead of $\vx^{i}$, $j \neq i$, is large (at least $\mu$).
Look at
Figure~\ref{fig:setmargin}, where, for simplicity, we need to choose a
pair ($k=2$).  Eq. \ref{eq:wxconstr} is represented by a
red line, that partitions the space of feasible utility weights in two
parts (in general, there will be $k$ subregions).  Since we maximize
the margin $\mu$, the optimizer will prefer a set of configurations
$\{ \vx^{1}, \vx^{2} \}$ that partitions the weight space in an ``even''
way.\footnote{This bears similarity with volumetric approaches \cite{iyengar:acm-ec01}, but there are important differences: first here
we consider real items to find the best separator, second the margin is also expressed in utility terms, third the query is found via an optimization process.}
In each subregion, we have corresponding %utility weight vector
$\vw^{i}$ lying ``close'' to its centre. 
If, for example, the user indicates a
preference for $\vx^{1}$ over $\vx^{2}$, the feasible region %for the weights 
will then become the part of the polytope to the left of the
red line; moreover the vector $\vw^{1}$ will maximize the margin in the classic ($k=1$) sense in the new feasible region.
%Therefore, setwise maxmargin can be thought as a ``look-ahead'' model with respect to a standard ($k=1$) max-margin learner.
%Therefore, setwise maxmargin can be thought as the set that, given any possible preference about the items, ensure that the reduction of utility uncertainty is the greatest
%\paolo{ho fatto un tentativo di migliorare l'intuizione, ma e' ancora possibile migliorare...}

\paragraph{MILP Formulation.} This initial formulation is problematic
to solve, as Eq.~(\ref{eq:wxconstr}) involves quadratic terms over mixed
continuous integer variables. However, the problem can be reformulated as a
mixed integer linear program (MILP) by a suitable transformation. This
technique is rather common in operational research, see
e.g.~\cite{boutilier2006}.

Our goal is to replace Eq.~(\ref{eq:wxconstr}) with a set of linear
constraints. In order to do so, we introduce a set of fresh variables
$p^{i,j}_z$ for every $i,j\in[k]$ and $z\in[m]$. Assuming for the time
being that the new variables do satisfy the equation
$p^{i,j}_z = w^i_z x^j_z$, we rewrite the fourth component of the
objective function in terms of the new variables as:
$$ \gamma \sum_{i=1}^k \sum_{z=1}^m p^{i,i}_z$$
and, similarly, Eq.~(\ref{eq:wxconstr}) as:
\[ \forall \; i, j \in [k], i \neq j \;.\; \sum_{z=1}^m p^{i,i}_z - p^{i,j}_z \ge \mu \label{eq:pxconstr} \]
The fact that $p^{i,j}_z = w^{i}_z x^{j}_z$ is achieved by
setting the following additional constraints. We distinguish between two cases:
(i) $p^{i,i}_z$ and (ii) $p^{i,j}_z$ for $i \ne j$.  Recall that we are
maximizing the margin $\mu$. Now, due to Eq.~(\ref{eq:pxconstr}), the optimizer will
try to keep $p^{i,i}_z$ as large as possible and $p^{i,j}_z$ as small as
possible.

(Case i) We add an explicit upper bound:
$ p^{i,i}_z \le \min \{ w_\text{max} x^{i}_z, w^{i}_z \} $,
where $w_\text{max}$ is a sufficiently large constant.
On one hand, if $x^i_z = 0$ the product $w^i_z x^i_z$ evaluates to $0$, and so does
the upper bound $w_\text{max} x^{i}_z = 0$. On the other hand, if $x^i_z=1$
then the product $w^i_z x^i_z$ amounts to $w^i_z$, while the upper
bound reduces to $\min \{ w_\text{max}, w^{i}_z \}$. By taking a sufficiently
large constant $w_\text{max}$ (e.g. $w_\text{max} := \max_z w^\top_z$) the
upper bound simplifies to $w^i_z$. Since $p^{i,i}_z$ is being maximized, in
both cases it will attain the upper bound, and thus satisfy $p^{i,j}_z = w^i_z x^i_z$.

(Case ii) We add an explicit lower bound:
$ p^{i,j}_z \ge \max \{ 0, w^{i}_z - w_\text{max}(1 - x^{j}_z) \} $.
If $x^j_z = 1$ the lower bound simplifies to
$\max \{ 0, w^{i}_z \} = w^{i}_z$, due to the non-negativity of
$w^i_z$. Otherwise, if $x^j_z = 0$ then the lower bound becomes
$\max \{ 0, w^{i}_z - w_\text{max} \}$, where the second term is at
most $0$. Since $p^{i,j}_z$ is being minimized, in both cases it will
attain the lower bound, and thus satisfy
$p^{i,j}_z = w^i_z x^j_z$.\footnote{Since $\mu$ is upper-bounded by
Eq.~(\ref{eq:wyconstr}), in some cases the $p^{i,j}_z$ variables do not attain
the lower bound. As a consequence, the MILP reformulation of
Eq.~(\ref{eq:wxconstr}) is a (tight) approximation of the original one. This
has no impact on the quality of the solutions.}

% Substituting the above MILP constraints into the original non-linear
% formulation,
We thus obtain the following mixed-integer linear problem:
{\footnotesize
\begin{align}
    \max
        & \;\; \mu - \alpha \sum_{i=1}^k \| \veps^{i} \|_1 - \beta \sum_{i=1}^k \| \vw^{i} \|_1 + \gamma \sum_{i=1}^k \sum_{z=1}^m p^{i,i}_z
        \nonumber
    \\
    \text{s.t.}
        & \;\; \forall \; i \in [k], \forall \; h \in [n] \quad \langle \vw^{i}, \vy^{h}_+ - \vy^{h}_- \rangle \ge \mu - \varepsilon^{i}_h \nonumber
    \\
        & \;\; \forall \; i, j \in [k], i \neq j \quad \sum_{z=1}^m p^{i,i}_z - p^{i,j}_z \ge \mu
    \\
    %     & \;\; \forall \; i \in [k], \forall \; z \in [m] \nonumber
    % \\
        & \;\; \forall \; i, j \in [k], i \neq j, \forall \; z \in [m] \nonumber
    \\
        & \;\; \qquad p^{i,i}_z \le \min \{ w_\text{max} x^{i}_z, \, w^{i}_z \} \label{eq:volatile1}
    \\
        & \;\; \qquad p^{i,j}_z \ge \max \{ 0, \, w^{i}_z - w_\text{max}(1 - x^{j}_z) \} \label{eq:volatile2}
    \\
        & \;\; \forall \; i \in [k] \quad \vw^\bot \le \vw^{i} \le \vw^\top \label{eq:wbounds2}
    \\
        & \;\; \forall \; i \in [k] \quad \vx^{i} \in \calX_{\text{feasible}} \;, \veps^{i} \ge 0 \nonumber
%    \\
%        & \;\; \forall \; i \in [k] \quad \vx^{i} \in \calX_{\text{feasible}} \nonumber
%    \\
%        & \;\; \forall \; i \in [k] \quad \veps^{i} \ge 0 \nonumber
%    \\
%        & \;\; \mu \ge 0 \nonumber
\end{align}
}
which can be solved by any suitable MILP solver.

\paragraph{Set-wise max-margin.} The full {\sc SetMargin} algorithm
follows the usual preference elicitation loop. Starting from an
initially empty set of user responses $\calD$, it repeatedly solves
the MILP problem above using $\calD$ to enforce ranking
constraints on the weight vectors $\{\vw^i\}$. The generated
configurations $\{\vx^i\}$, which are chosen to be as good as possible
with respect to the estimated user preferences, and as diverse as
possible, are then employed to formulate a set of user queries. The
new replies are added to $\calD$ and the whole procedure is
repeated. Termination can be after a fixed number of iterations,
when the difference between utility vectors is very small, or might be left
to the user to decide (e.g. \cite{Reilly2007}).

The procedure is sketched in
Algorithm~\ref{alg:setmargin}. Note that at the end of the preference
elicitation procedure, a final recommendation is made by solving the
MILP problem for $k=1$.

\begin{algorithm}[t]
{\footnotesize
\begin{algorithmic}[1]
    \Procedure{SetMargin}{$k, \alpha, \beta, \gamma, T$}
        \State $\calD \gets \emptyset$
        \For{$t = 1, \ldots, T$}
            \State \{$\vw^{i}, \vx^{i}\}_{i=1}^k \gets \text{{\sc Solve}}(\calD, k, \alpha, \beta, \gamma)$
            \For{$\vx^{i},\vx^{j} \in \{ \vx^{1}, \ldots, \vx^{k} \} \; \text{{\bf s.t.}} \; i < j$}
                \State $\calD \gets \calD \cup \text{{\sc QueryUser}}(\vx^{i},\vx^{j})$
            \EndFor
        \EndFor
        \State $\vw^*, \vx^* \gets \text{{\sc Solve}}(\calD, 1, \alpha, \beta, \gamma)$
        \State ${\bf return}\; \vw^*, \vx^*$
    \EndProcedure
\end{algorithmic}
}
\caption{\label{alg:setmargin} The {\sc SetMargin} algorithm. Here $k$ is the
set size, $\alpha,\beta,\gamma$ are the hyperparameters, and $T$ is the maximum
number of iterations. The values of $\calX_\text{feasible}$, $\vw^\top$ and
$\vw^\bot$ are left implicit.}
\end{algorithm}

\paragraph{Linearly dependent real attributes.} In many domains of
interest, items are composed of both Boolean and real-valued
attributes, where the latter depend linearly on the former. This is
for instance the case for the price, weight and power consumption of a
laptop, which depend linearly on the choice of components.  In
this setting, %item 
configurations are composed of two parts:
$\vx = (\vx_B;\vx_R)$, where $\vx_B$ is Boolean and $\vx_R$ is
real-valued and can be written as $\vx_R = C \vx_B$ for an
appropriately sized non-negative cost matrix $C$.  It is
straightforward to extend the MILP formulation to this setting. %First,
We rewrite the weight vector as $\vw = (\vw_B;\vw_R)$. The utility
becomes:
$$ \langle \vw, \vx \rangle = \langle \vw_B, \vx_B \rangle + \langle \vw_R, C \vx_B \rangle = \langle \vw_B + C' \vw_R, \vx_B \rangle $$
The generalized problem is obtained by substituting $\vw^i$ with $\vv^i :=
\vw_B^i + C' \vw_R^i$.  All constraints remain the same. The only notable
change occurs in Eq.~(\ref{eq:wbounds2}), which becomes:
$$ \forall \; i \in [k] \;.\; (\vw_B^\bot + C' \vw_R^\bot) \le \vv^i \le (\vw_B^\top + C' \vw_R^\top)$$
%
%As can be seen, non-negativity of $C$ is required for the weights $\{\vv^i\}$
%to be non-negative.

\section{Experiments}
\label{sec:experiments}

We implemented the {\sc SetMargin} algorithm using Python, leveraging Gurobi
6.5.0 for solving the core MILP problem. Both the {\sc SetMargin} source code
and the full experimental setup are available at 
\url{https://github.com/stefanoteso/setmargin}.

We compare {\sc SetMargin} against three state-of-the-art Bayesian
approaches: i) the Bayesian approach from \cite{guo2010real},
selecting queries according to {\em restricted informed VOI} ({\sc
  riVOI}), a computationally efficient heuristic approximation of
value-of-information, and inference using TrueSkill$^{TM}$
\cite{HerbrichMG06} (based on expectation
propagation \cite{Minka01}); ii) the Bayesian framework of
\cite{viappiani2010optimal} using Monte Carlo methods (with 50,000
particles) for Bayesian inference and asking choice queries
(i.e. selection of the most preferred item in a set) selected
using a greedy optimization of {\em Expected Utility of a Selection} (a tight
approximation of EVOI, hereafter just called {\sc
  EUS}); iii) {\em Query Iteration} (referred as {\sc QI} below), also
from \cite{viappiani2010optimal}, an even faster query selection
method based on sampling sets of utility vectors. % that often converge to query set of high value of information.

We adopt the {\em indifference-augmented} Bradley-Terry user response
model introduced in~\cite{guo2010real}. The probability that a user
prefers configuration $\vx^i$ over $\vx^j$ is defined according to the
classical (without indifference) Bradley-Terry model~\cite{BraTer52} as
$ (1 + \exp(-\lambda_1 \langle\vw,\vx^i - \vx^j\rangle))^{-1} $,
where $\vw$ is the weight vector of the true underlying user utility.
Support for indifference is modelled as an exponential distribution
over the closeness of the two utilities, i.e. 
$ \exp(-\lambda_2 |\langle\vw,\vx^i - \vx^j\rangle|).$
The parameters $\lambda_1$ and $\lambda_2$ were set to one for all
simulations, as in~\cite{guo2010real}.
% \footnote{When experimenting with EUS and QI, the user is
%   forced to select its preffered item, so indifference is not
%   possible. While support for indifference could be added, we chose to
%   be consistent with the methods as originally published.}, as

In all experiments {\sc SetMargin} uses an internal 5-fold cross-validation procedure to
update the hyperparameters $\alpha$, $\beta$, and $\gamma$ after every 5
iterations. The hyperparameters are chosen as to minimize the ranking loss over
the user responses collected so far. $\alpha$ is taken in $\{20, 10, 5, 1\}$,
while $\beta$ and $\gamma$ are taken in $\{10, 1, 0.1, 0.001\}$.\footnote{Note
that for $k=1$ Eq.~(\ref{eq:volatile1}) and Eq.~(\ref{eq:volatile2}) disappear,
so $\alpha$ can not be taken to be less than $1$, as in this case, the
objective can be increased arbitrarily while keeping the right-hand side of
Eq.~(\ref{eq:wyconstr}) constant, rendering the problem unbounded.}

\begin{figure*}[t]
    \centering
    {\footnotesize
    \begin{tabular}{ccccc}
        % \hline
        % {\sc synthetic 3 loss} & {\sc synthetic 3 time} & {\sc synthetic 4 loss} & {\sc synthetic 4 time}
        % \\
        % \hline \hline
        % \multicolumn{4}{c}{{\sc Uniform}}
        \\
        {\sc Uniform}  &
        \includegraphics[align=c,width=10em]{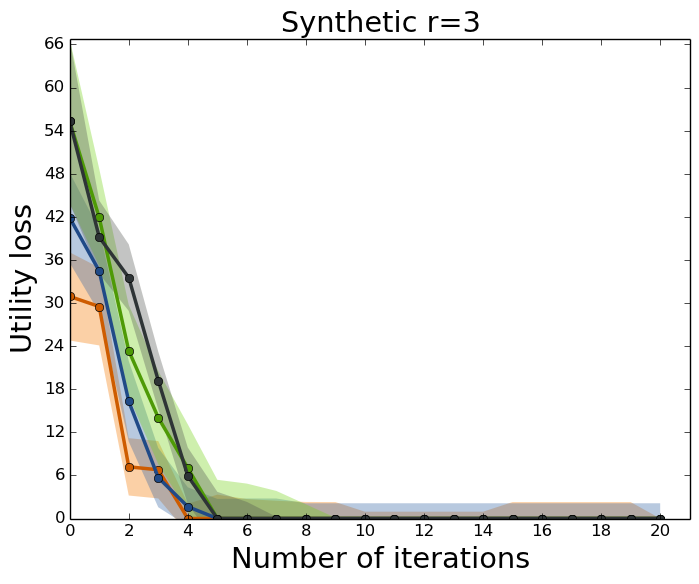} &
        \includegraphics[align=c,width=10em]{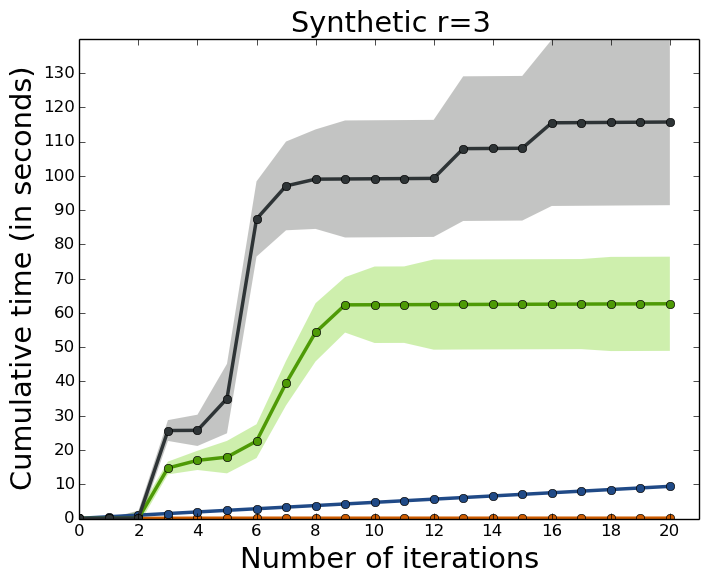} &
        \includegraphics[align=c,width=10em]{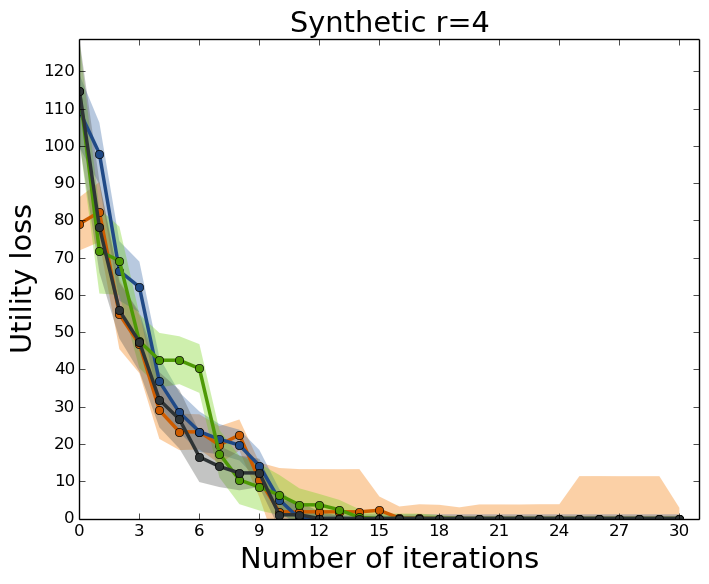} &
        \includegraphics[align=c,width=10em]{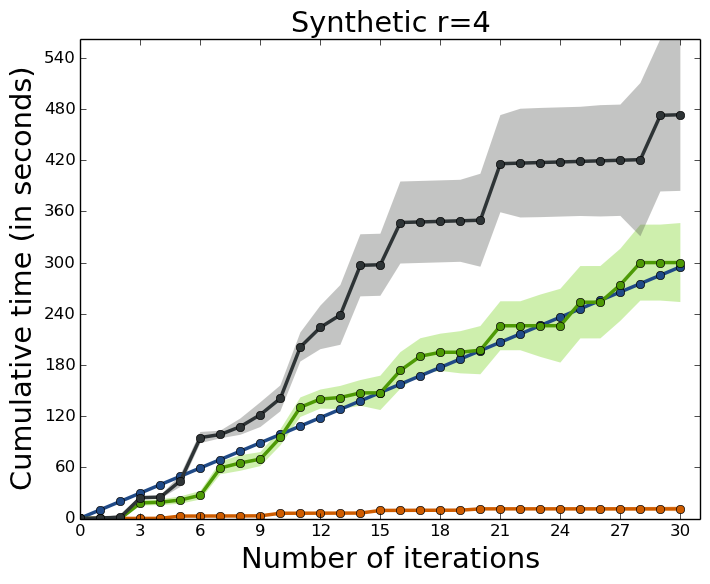}
        \\
        \hline
         % \multicolumn{4}{c}{{\sc Normal}}
        \\
         {\sc Normal} &
        \includegraphics[align=c,width=10em]{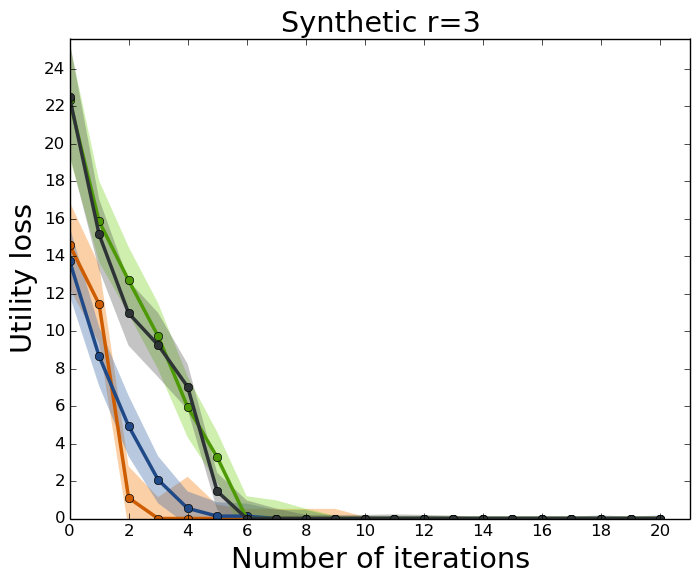} &
        \includegraphics[align=c,width=10em]{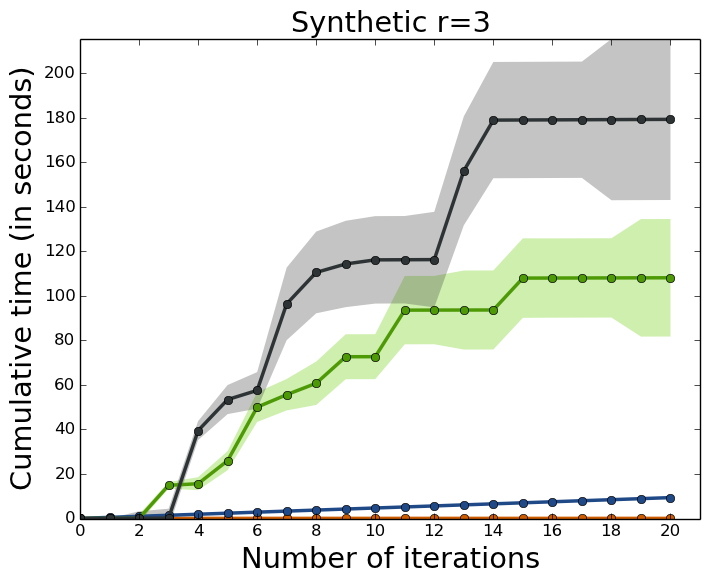} &
        \includegraphics[align=c,width=10em]{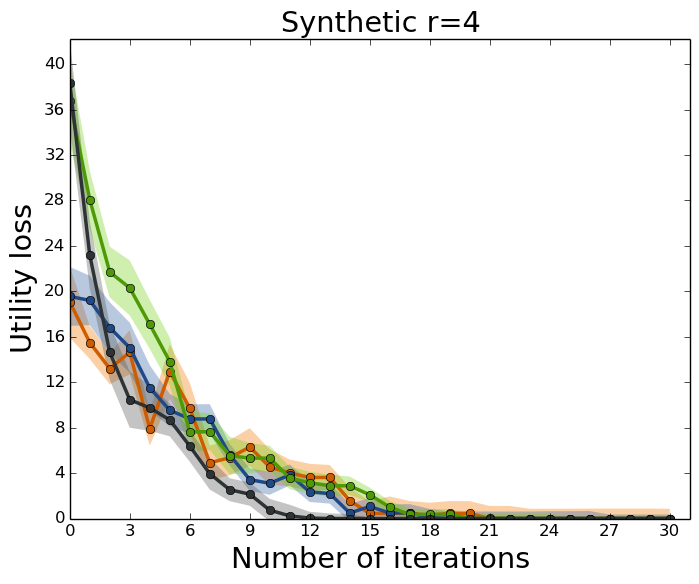} &
        \includegraphics[align=c,width=10em]{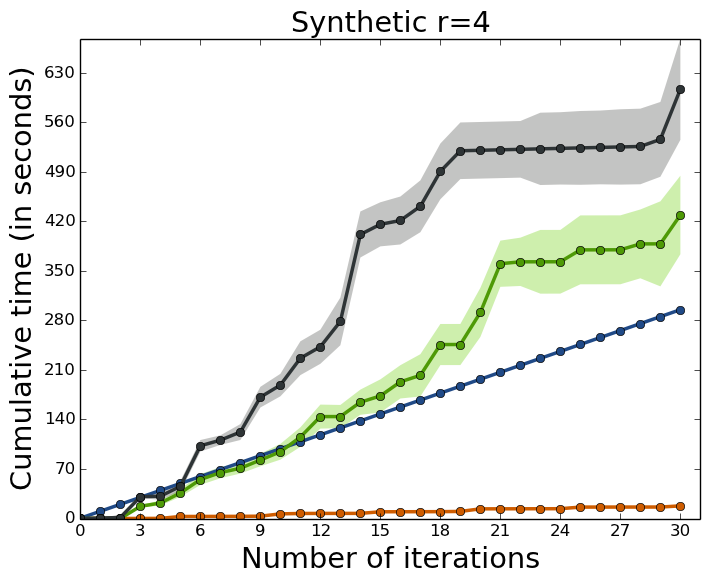}
        \\
        \hline
        % \multicolumn{4}{c}{{\sc Sparse Uniform}}
        \\
        {\sc Sparse Uniform} &
        \includegraphics[align=c,width=10em]{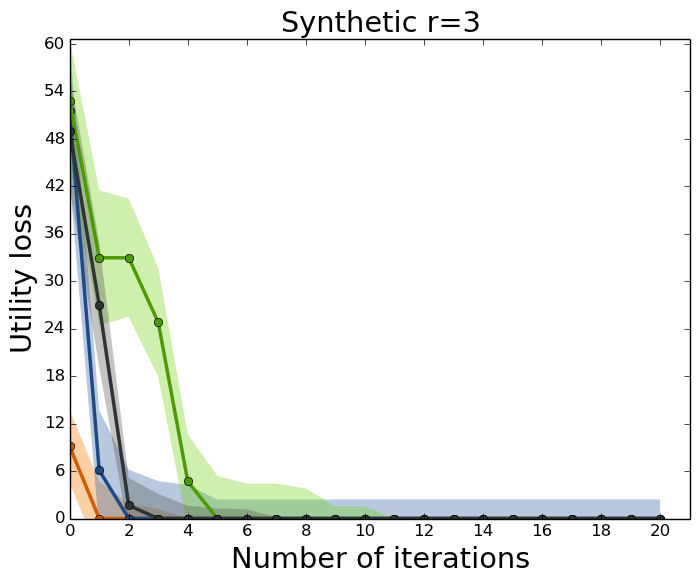} &
        \includegraphics[align=c,width=10em]{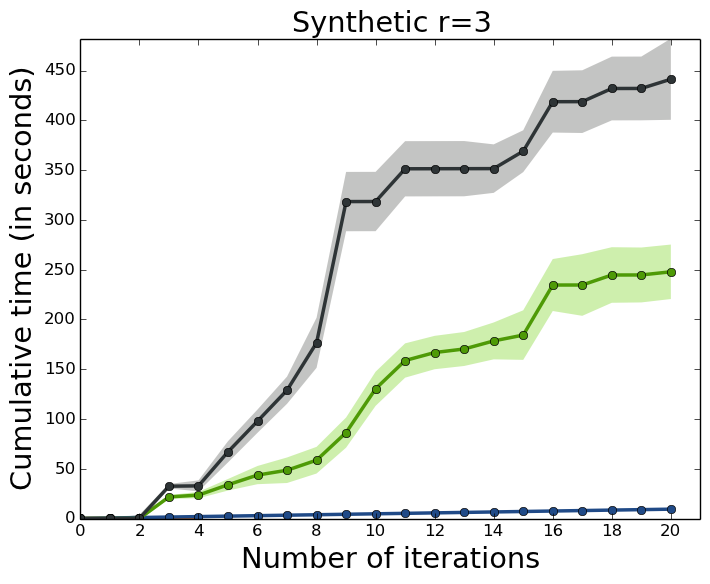} &
        \includegraphics[align=c,width=10em]{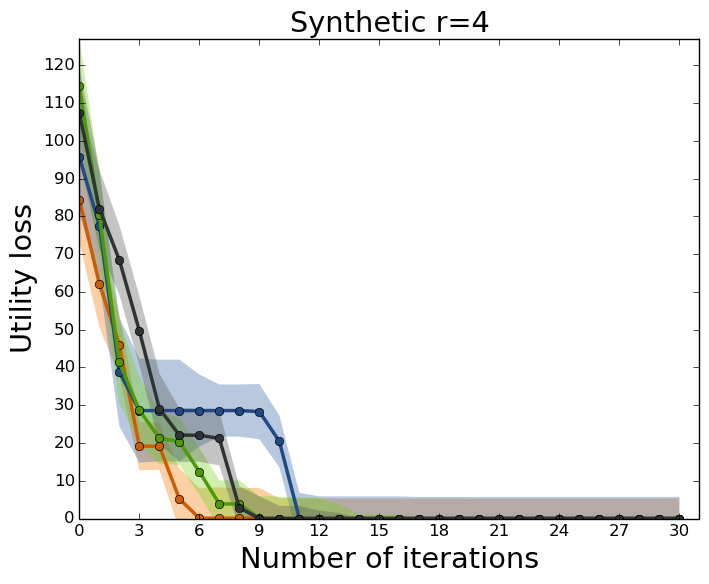} &
        \includegraphics[align=c,width=10em]{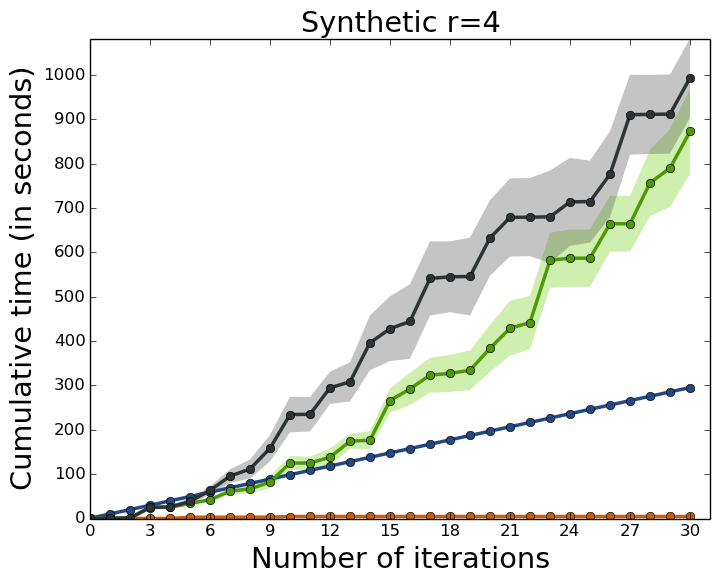}
        \\
        \hline
        % \multicolumn{4}{c}{{\sc Sparse Normal}}
        \\
        {\sc Sparse Normal} &
        \includegraphics[align=c,width=10em]{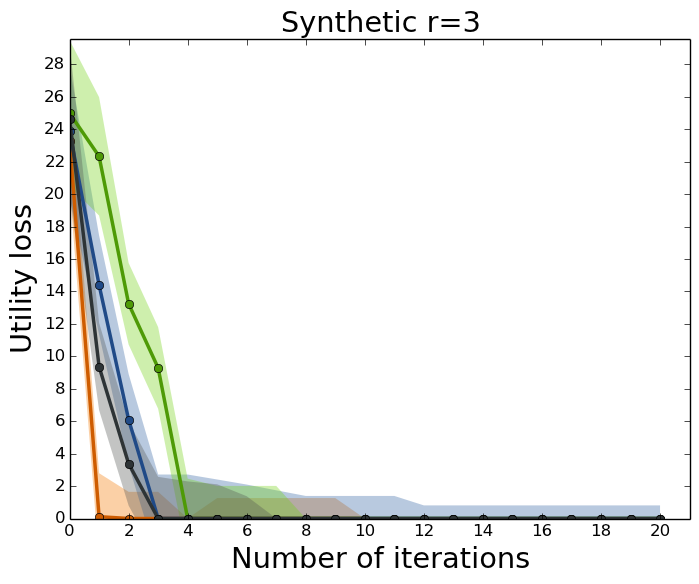} &
        \includegraphics[align=c,width=10em]{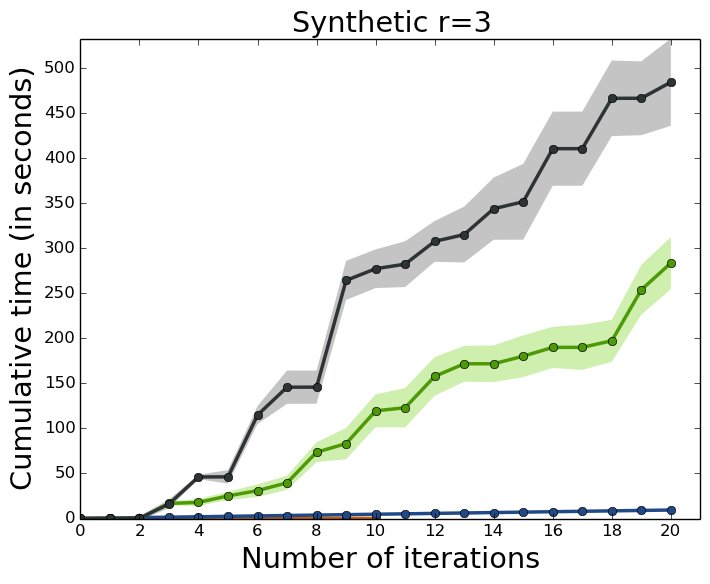} &
        \includegraphics[align=c,width=10em]{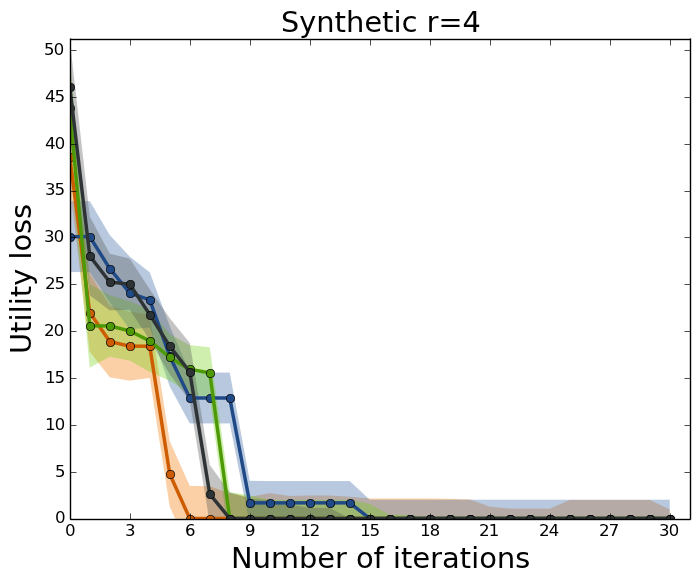} &
        \includegraphics[align=c,width=10em]{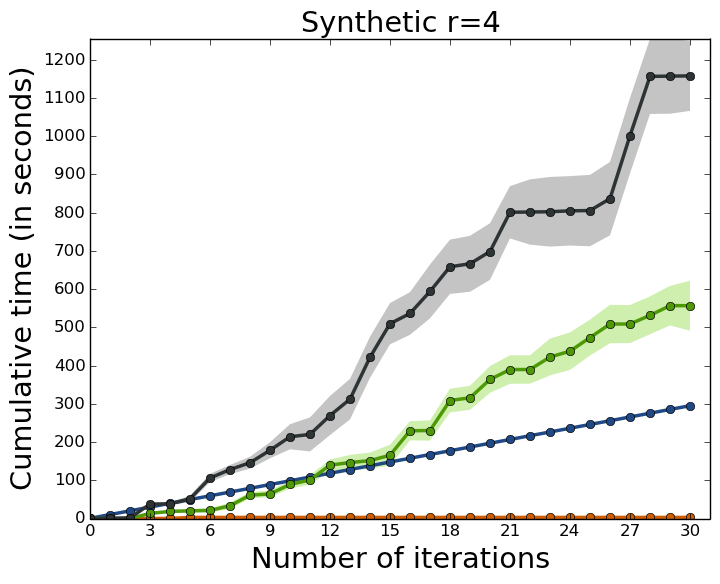}
        \\
        %\hline
    \end{tabular}
    }
    \caption{\label{fig:comparison} Comparison between {\sc SetMargin}
      (orange), {\sc riVOI} (blue), {\sc QI} (green) and {\sc EUS}
      (gray) on the $r=3$ (left) and $r=4$ (right) datasets. Each row
      represents a different sampling distribution for user utility.
      The number of iterations is plotted against the utility loss
      (first and third columns) and the cumulative time (second and
      fourth columns). Thick lines indicate median values over users,
      while standard deviations are shown as shaded areas.}
\end{figure*}

\paragraph{Synthetic Dataset.} Following the experimental protocol
in~\cite{guo2010real} and \cite{viappiani2010optimal}, in the first
experiment we evaluate the behavior of the proposed method in an
artificial setting with increasingly complex problems. We developed
synthetic datasets with $r$ attributes, for increasing values of $r$.
Each attribute takes one of $r$ possible values, so that the one-hot
encoding of attributes results in $m=r^2$ features. In terms of space
of configurations, for $r=3$ the synthetic dataset corresponds to 
$\calX_\text{feasible} = [3] \times [3] \times [3]$, for $r=4$ to
$\calX_\text{feasible} = [4] \times [4] \times [4] \times [4]$, and so on. The
cardinality of $\calX_\text{feasible}$ is $r^r$, and grows (super)
exponentially with $r$.
% We developed
% seven synthetic datasets: each dataset involves $m=2,\ldots,7$
% attributes, where each attribute takes one of $m$ possible
% values. More explicitly, for $m=2$ the synthetic dataset is
% $\calX_\text{feasible} = [2] \times [2]$, for $m=3$ to
% $\calX_\text{feasible} = [3] \times [3] \times [3]$, and so on. The
% cardinality of $\calX_\text{feasible}$ is $m^m$, and grows (super)
% exponentially with $m$. 
For $r=3$, the dataset is comparable in size to the synthetic one used
in~\cite{guo2010real} and~\cite{viappiani2010optimal}.
%, which have
%three attributes with 2, 2 and 5 values respectively, for a total of
%20 feasible configurations.  
For larger $r$ the size of the space grows much larger than the ones
typically used in the Bayesian preference elicitation literature, and
as such represents a good testbed for comparing the scalability of the
various methods. The feasible configuration space was encoded in {\sc
  SetMargin} through appropriate MILP constraints, while the other
methods require all datasets to be explicitly grounded.  Users were
simulated by drawing $20$ random utility vectors from each of four
different distributions. The first two mimic those used
in~\cite{guo2010real}
%\footnote{We used the parameters implemented in their code, which are slightly different from those reported in the paper.}%
%\footnote{The normal distribution reported
%  in~\cite{guo2010real} differs slightly from the one implemented in
%  the code of their experimental setting.  In our experiments we use
%  the latter.}
: (1) a uniform distribution over $[1, 100]$ for each
individual weight, and (2) a normal distribution with mean $25$
and %covariance
standard deviation $\frac{25}{3}$ (each attribute is sampled i.i.d).
We further produced two novel {\em sparse} versions of the uniform and
normal distributions setting to zero $80\%$ of the entries (sampled
uniformly at random). We set a maximum budget of 100 iterations for all
methods for simplicity.

% ; for {\sc SetMargin}a
% we used a $0.01$ threshold on the utility loss for early stopping
% (that does not affect the cross-validation procedure).  The Bayesian
% methods can use the estimated loss to decide when to stop (the
% expectation of the utility loss with respect to the current
% probabilistic belief) \footnote{In our experiments, we left the
%   Bayesian methods to run the full amount of iterations, even when the
%   estimated loss was very low.}. \andrea{@stefano: cosi' non va. O tutti early stopping o nessuno..}

In Figure~\ref{fig:comparison} we report solution quality and timing
values for increasing number of collected user responses, for the
different competitors on each of the four different utility vector
distributions and datasets $r\!=\!3$ and $r\!=\!4$. Solution quality is
measured in terms of utility loss $ \max_{\vx\in\calX_\text{feasible}} \left( u(\vx) - u(\vx^*) \right)$,
where $u(\cdot)$ is the true unknown user utility, and $\vx^*$ is the
solution recommended to the user after the elicitation phase (see
Algorithm~\ref{alg:setmargin}). Computational cost is measured in
terms of cumulative time. Given that {\sc riVOI},  {\sc QI} and
{\sc EUS} are single-threaded, we disabled multi-threading when running our
algorithm in these comparisons. All experiments were run on a 2.8 GHz Intel
Xeon CPU with 8 cores and 32 GiB of RAM.
%In all figures, thick lines indicate median values over the different
%users, shaded areas indicate standard deviations. Our {\sc SetMargin}
%algorithm for $k=2$ is reported in orange, {\sc riVOI} in blue, {\sc QI}
%in green and {\sc EUS} in gray. 
For all algorithms, one iteration corresponds to a single pairwise query (we
used {\sc SetMargin} with $k=2$).
%query to the user 
%Results for $r=3$ and $r=4$ are reported.  
For dense weight vector distributions (first two rows), our approach
achieves results which are indistinguishable from the competitors in a
fraction of their time.  Indeed, all Bayesian approaches
%\footnote{Note
%  that for {\sc QI} and {\sc EUS}, the time spent for Bayesian
%  inference greatly dominates the time spent for choosing the next
%  query to ask.\andrea{@paolo: necessario?}} 
 become quickly impractical for growing values of $r$,
%cannot scale over $r=4$ (for $r=5$, {\sc riVOI} did not
%finish a single user after 24 hours for, Viappiani QI takes 20 minutes
%per user), \andrea{20 minuti mi paiono pochi, sei sicuro?}  
while our algorithm can easily scale to much larger datasets, as will
be shown later on. For sparse weight vector distributions (last two
rows) our approach, in addition to being substantially faster on each
iteration, requires less queries in order to reach optimal
solutions. This is an expected result as the sparsification norm in
our formulation ($\| \vw \|_1$) is enforcing sparsity in the weights, 
while none of the other approaches is %explicitly 
designed to do this. %favour sparse functions.

% {\sc riVOI} e' piantato al primo utente da tre giorni su synthetic 5.
% Paolo procede su S-5 da qualche ora (~8) (1290s su synthetic-5
% uniforme per completamento di un utente).

%\stefano{@Paolo: varianza tempi?}

\begin{figure*}[t]
    \centering
    {\footnotesize
    \begin{tabular}{ccccc}
        % \hline
        % \multicolumn{2}{c}{{\sc synthetic 4}} &
        % \multicolumn{2}{c}{{\sc synthetic 5}}
        % \\
        % {\sc loss per iter.} & {\sc loss per query} & {\sc loss per iter.} & {\sc loss per query}
        % \\
        % \hline \hline
        {\sc uniform} &
        \includegraphics[align=c,width=10em]{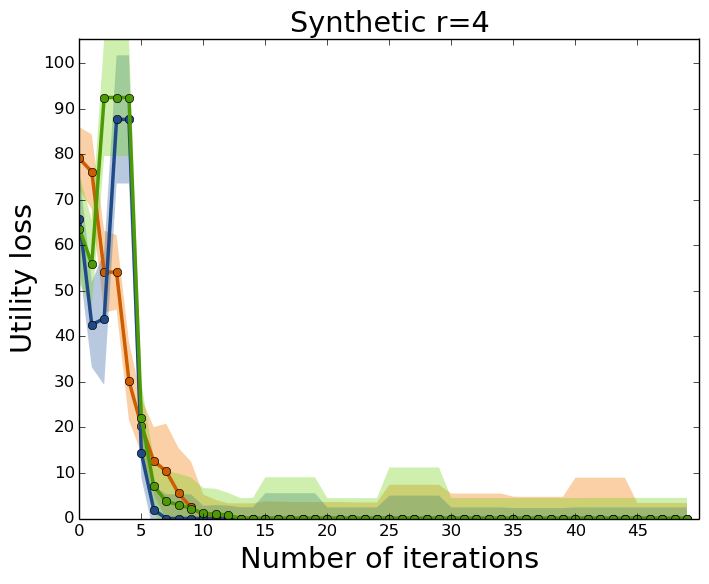} &
        \includegraphics[align=c,width=10em]{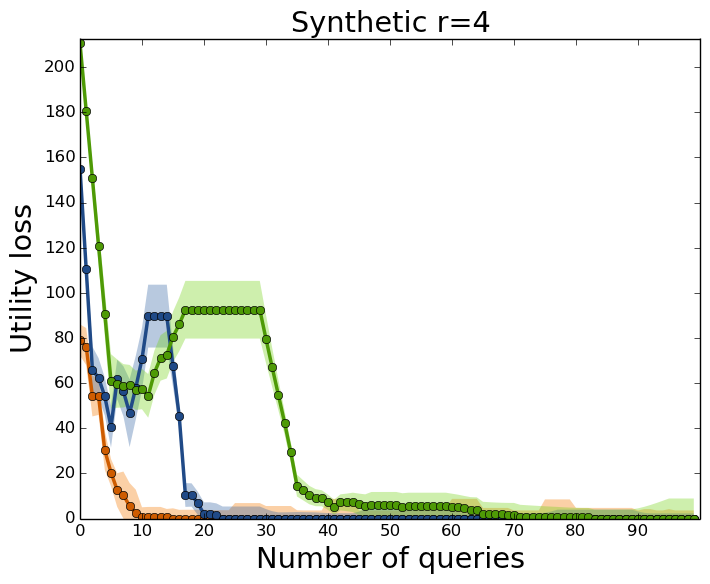} &
        \includegraphics[align=c,width=10em]{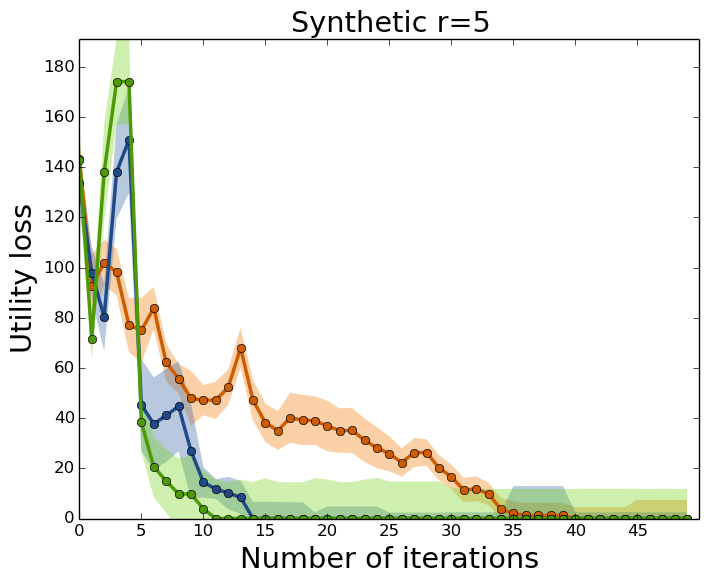} &
        \includegraphics[align=c,width=10em]{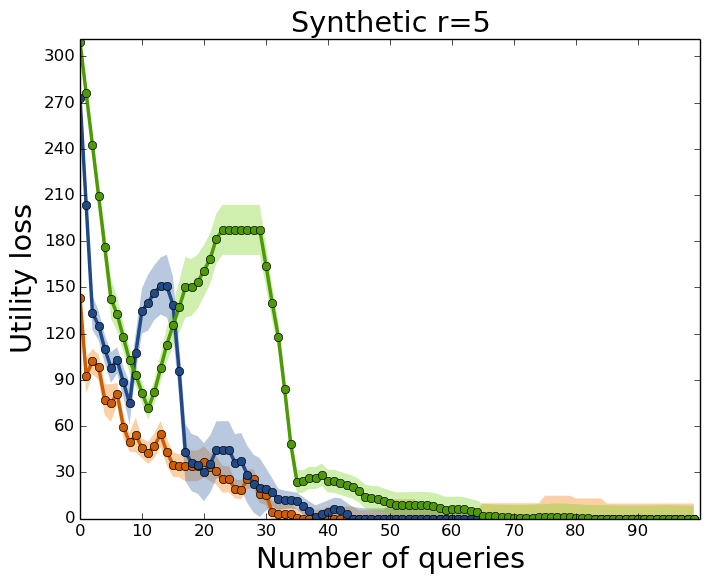}
        \\
        \hline
        {\sc sparse normal} &        
        \includegraphics[align=c,width=10em]{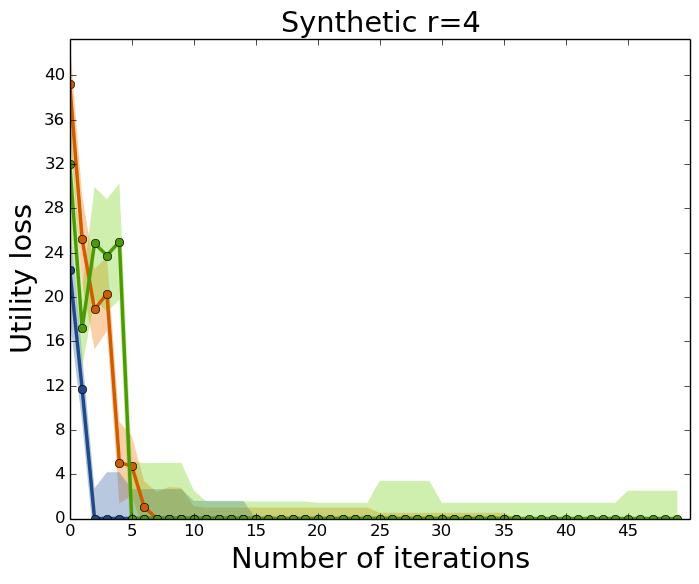} &
        \includegraphics[align=c,width=10em]{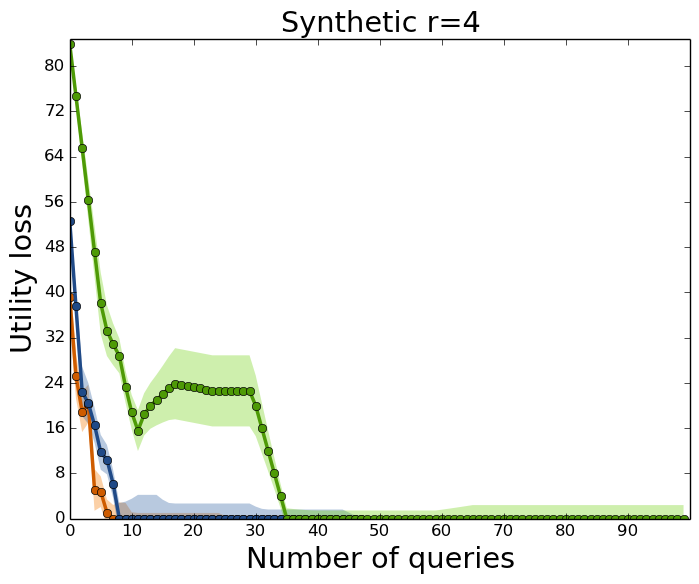} &
        \includegraphics[align=c,width=10em]{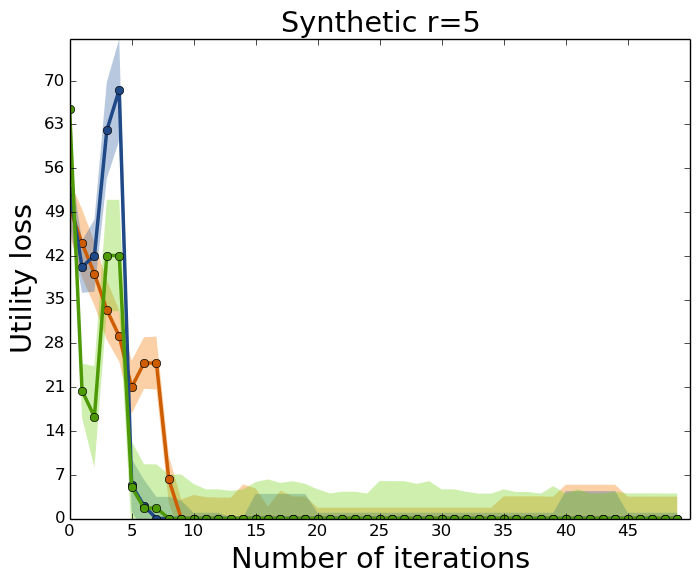} &
        \includegraphics[align=c,width=10em]{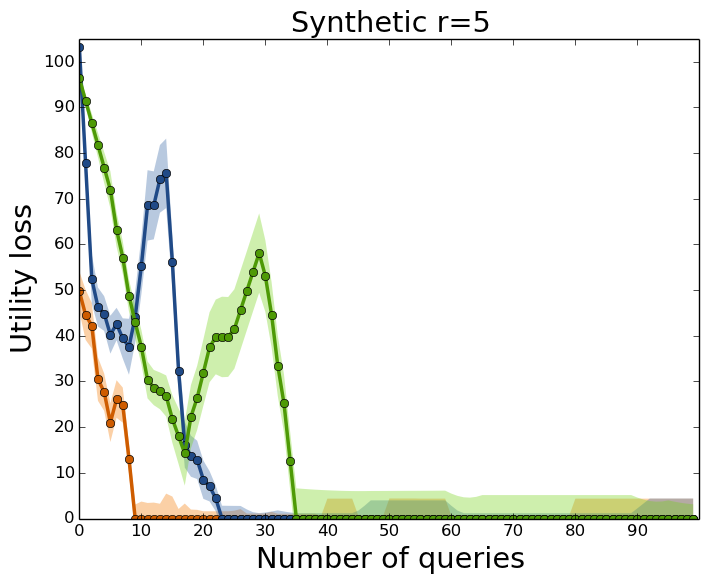}
        \\
        %\hline
     \end{tabular}
    }
    \caption{\label{fig:selfcomparison} Comparison for {\sc SetMargin}
      with $k=2,3,4$, in orange, blue and green respectively for the
      $r=4$ (left) and $r=5$ (right) datasets using uniform (top row) and sparse normal (bottom row) distributions. Median and standard deviation utility loss values are reported for increasing number of iterations ($1^{st}$ and $3^{rd}$ columns) and pairwise queries ($2^{nd}$ and $4^{th}$ columns).}
\end{figure*}

\begin{figure*}
    \centering
    {\footnotesize
    \begin{tabular}{lcccc}
%        \hline
%        \multicolumn{2}{c}{{\sc loss/time per iter.}} &
%        \multicolumn{2}{c}{{\sc loss/time per query}}
        \\
%        \multicolumn{4}{c}{{\sc sparse uniform}}
        {\sc sparse uniform} &
        \includegraphics[align=c,width=10em]{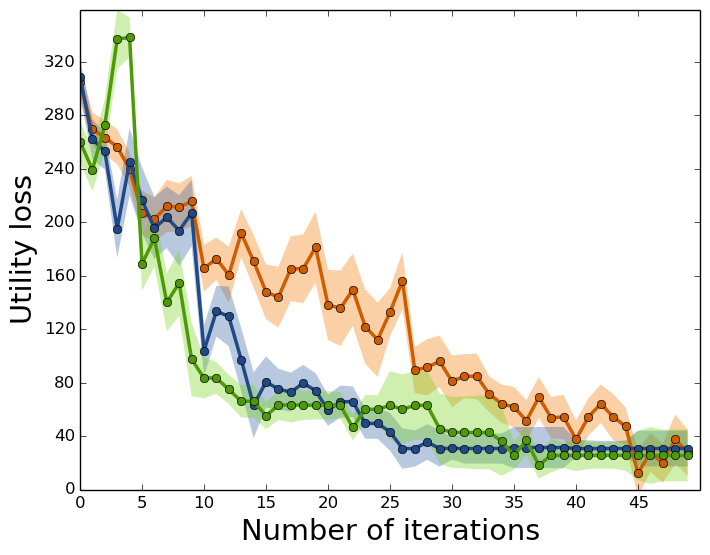} &
        \includegraphics[align=c,width=10em]{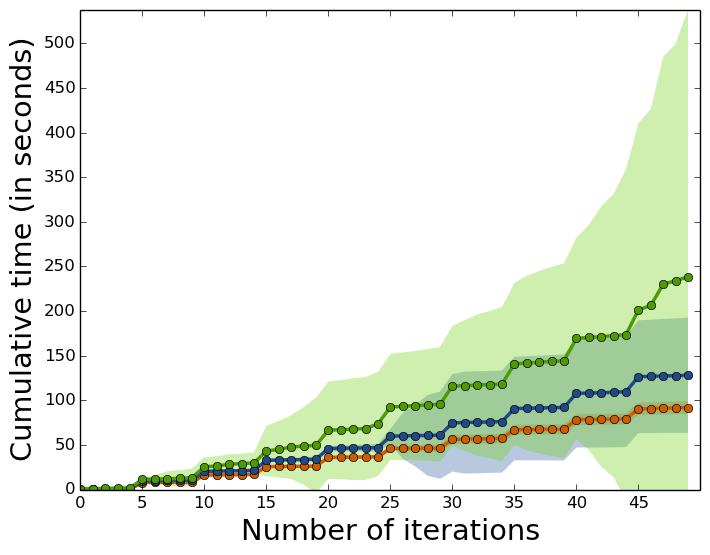} &
        \includegraphics[align=c,width=10em]{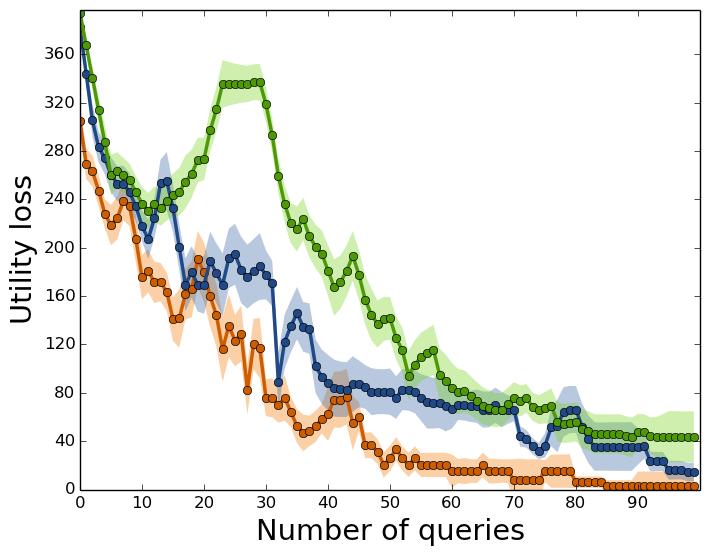} &
        \includegraphics[align=c,width=10em]{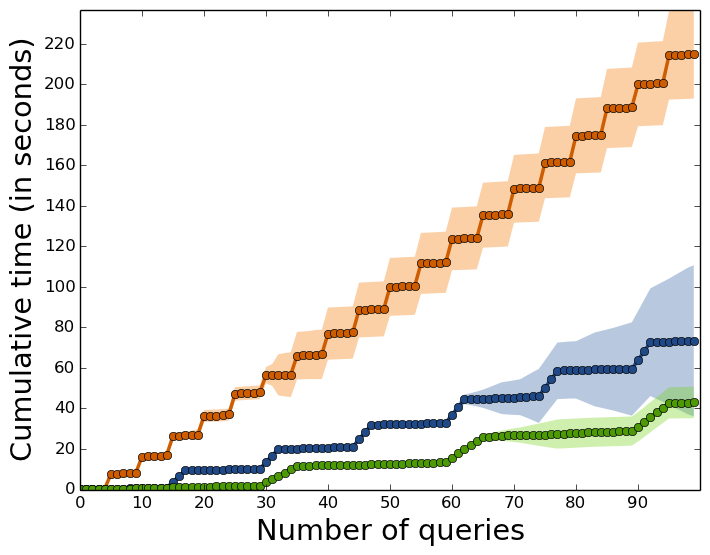}

        \\
%        \hline
        % \multicolumn{4}{c}{{\sc sparse normal}}
        % \\
        % \includegraphics[width=10em]{figures/pc_with_costs_normal_sparse_per_iter_loss} &
        % \includegraphics[width=10em]{figures/pc_with_costs_normal_sparse_per_iter_time} &
        % \includegraphics[width=10em]{figures/pc_with_costs_normal_sparse_per_query_loss} &
        % \includegraphics[width=10em]{figures/pc_with_costs_normal_sparse_per_query_time}
        % \\
        % \hline
    \end{tabular}
    }
    \caption{\label{fig:pc} Results for {\sc SetMargin} for $k=2$
      (red), $k=3$ (blue) and $k=4$ (green) on the constructive PC
      dataset for the sparse uniform distribution. Utility loss and
      time are plotted against number of iterations (left)
      and number of queries (right).}
\end{figure*}

In order to study the effect of increasing the number of weight
vectors in our formulation, we also ran {\sc SetMargin} varying the
parameter $k$. Figure~\ref{fig:selfcomparison} reports utility loss
results on $r\!=\!4$ and $r\!=\!5$ datasets for the uniform and sparse normal
distributions (the toughest and the simplest, for space limitations).
% \footnote{We picked the most difficult dense distribution
%   (uniform) and the simplest sparse distribution (normal) to provide
%   the broadest possible picture, as space constraints prevent us from
%   including all results. Results for the other distributions are
%   however qualitative similar to the ones reported.}, 
The first and third columns report results in terms of number of
iterations. It can be seen that increasing the number of weight
vectors tends to favour earlier convergence, especially for the more
complex dataset ($r=5$). However, as in each iteration the user is
asked to compare $k$ items, different values of $k$ imply a different
cognitive effort for the user. The second and fourth columns report
results in terms of number of queries, where we count all
$\binom{k}{2}$
%k \choose 2$ 
pairs of queries when comparing $k$ items. In this case,
$k=2$ seems to be the best option. The cognitive cost for the user
will likely lay in between these two extremes, but formalizing this
concept in an efficient query ordering strategy needs to face the
effect of noise.  A modified sorting algorithm asking only
$O(k\log k)$ queries to the user resulted in a performance worsening,
likely because of a cascading effect of inconsistent feedback (but
could be beneficial with different noise levels).
% \paragraph{Cognitive effort.} Contrarily to most other preference elicitation
% methods, which generate a single candidate option per iteration, {\sc
% SetMargin} constructs a whole {\em set} of options. We thus can not leverage
% standard query selection strategies \stefano{which ones?}. 
% The simplest solution is to query the user about all $k \choose 2$ pairs of
% configurations in $\{\vx^i\}$. Since $k$ is a small constant, the resulting
% number of queries does not grow large in practice.  An appealing alternative
% involves using a modified sorting algorithm, such as merge-sort, to order the
% set $\{\vx^i\}$ by asking only $O(n\log n)$ queries to the user. However, this
% approach is very susceptible to noise: a single inconsistent user response may
% affect all successive comparisons, leading to a cascading effect. While more
% robust strategies may be conceived, empirically we found that the simpler
% quadratic scheme tends to work reasonably well.
%\paolo{Maybe a plot of an experiment with increased noise in user response? (for example, $\beta=0.1$ instead of $1$.}

\paragraph{Constructive dataset.} Next, we tested {\sc SetMargin} on a truly
constructive setting. We developed a constructive version of the PC dataset
used in~\cite{guo2010real}: instead of explicitly enumerating all possible PC
items, we defined the set of feasible configurations with MILP constraints.

A PC configuration is defined by eight attributes: computer type (laptop,
desktop, or tower), manufacturer (8 choices), CPU model (37), monitor size (8), RAM amount (10), storage (10)
size, and price.
%See Table~\ref{tab:pcdataset} for an overivew of the attributes.
The price attribute is defined as a linear combination of the other
attributes: this is a fair modeling choice, as often the price of a PC is
well approximated by the sum of the price of its components plus a bias due to
branding.
Interactions between attributes are expressed as Horn clauses (e.g. a certain manufacturer implies a set of possible CPUs).
% . An example
% constraint might look like this: ``if the manufacturer is Apple, then the
% CPU must be either a PowerPC G3 or a PowerPC G4''%
% %\footnote{Despite the components
% %included in the dataset, which are a bit outdated, the dataset itself is pretty
% %realistic.} \andrea{necessario?}
% , which can be encoded into MILP form as
% %
% $ (1 - x_\text{Apple}) + x_\text{G3} + x_\text{G4} \ge 1 $.
% %
% Note that only one of $x_\text{G3}$ and $x_\text{G4}$ can be 1 because
% of the mutual exclusivity constraints of the one-hot encoding.  
The
dataset includes 16 Horn constraints (the full list is omitted for
space limitations).
% constraints between the following attributes:
% manufacturer $\to$ type, manufacturer $\to$ CPU, type $\to$ RAM
% amount, type $\to$ storage size, type $\to$ monitor size, for 16 Horn
% constraints total. We do not report the full list due to space
% limitations. 
Note that the search space is of the order of hundreds of thousands of
candidate configurations, and is far beyond reach of existing Bayesian
approaches.

Figure~\ref{fig:pc} reports results of {\sc SetMargin} varying $k$
using the sparse uniform distribution (the more complex of the sparse
ones, dense distributions being unrealistic in this scenario). The
first and third column report utility loss for increasing number of
iterations and queries respectively, showing a behaviour which is
similar to the one in Figure~\ref{fig:selfcomparison}. Overall,
between 50 and 70 queries on average are needed in order to find  a
solution which is only 10\% worse than the optimal one, out of the
more than 700,000 thousands available. Note that a vendor may ensure a
considerably smaller number of queries by cleverly constraining the
feasible configuration space; since our primary aim is benchmarking,
we chose not to pursue this direction further.  The second and fourth
columns report cumulative times. Note that in some cases, standard
deviations have a bump; this is due to cases in which some of the
hyperparameters of the internal cross validation result in
ill-conditioned optimization problems which are hard to solve. These
exceptions can be easily dealt with by setting an appropriate timeout
on the cross validation without affecting the results, as these
hyperparameters typically end up having bad performance and being
discarded.

% \begin{table}
%     \centering
%     \begin{tabular}{ccc}
%         {\bf Attribute} & {\bf Type} & {\bf Values} \\
%         \hline \hline
%         Type & discrete & 3 \\
%         Manufacturer & discrete & 8 \\
%         CPU model & discrete & 37 \\
%         RAM amount & discrete & 10 \\
%         HD size & discrete & 10 \\
%         Monitor size & discrete & 8 \\
%         Price & continuous & --
%     \end{tabular}
%     \caption{\label{tab:pcdataset} Statistics for the PC dataset.}
% \end{table}

\section{Conclusion}
\label{sec:conclusions}

We presented a max-margin approach for efficient preference
elicitation in large configuration spaces.\footnote{Note that max-margin learning has been proposed before \cite{gajos2005} for preference elicitation, 
but with  rudimental methods for query selection.}
Our approach relies on an
extension of max-margin learning to sets, and is effective in the
generation of a diverse set of configurations that can be used to ask
informative preference queries.  The main advantages of this
elicitation method are 1) ability to provide recommendations in large
configuration problems 2) robustness with respect to erroneous
feedback and 3) ability to encourage sparse utility functions.
Experimental comparisons against state-of-the-art Bayesian preference
elicitation strategies confirm these advantages. Future work includes
extending the approach to truly hybrid scenarios (where real valued
attributes do not depend on categorical ones) and studying its
applicability to other problems, as the identification of Choquet models
\cite{Pine2013}.

%\section*{Acknowledgments}
\paragraph{Acknowledgments}
ST was supported by the Caritro Foundation through project E62I15000530007.
PV was supported by the Idex Sorbonne Universit\'{e}s under grant ANR-11-IDEX-0004-02.
We thank Craig Boutilier for motivating discussion on the topic.

\bibliographystyle{named}
\bibliography{ijcai16}

% times for the figure above
% synthetic_vs_self_4_normal_sparse_per_iter_time
% synthetic_vs_self_4_normal_sparse_per_query_time
% synthetic_vs_self_4_uniform_per_iter_time
% synthetic_vs_self_4_uniform_per_query_time
% synthetic_vs_self_5_normal_sparse_per_iter_time
% synthetic_vs_self_5_normal_sparse_per_query_time
% synthetic_vs_self_5_uniform_per_iter_time
% synthetic_vs_self_5_uniform_per_query_time

\end{document}